# Ethical Implications of ChatGPT in Higher Education: A Scoping Review

Ming Li
*Osaka University/Japan*
Ariunaa Enkhtur
*Osaka University/Japan*
Fei Cheng
*Kyoto University/Japan*
Beverley Anne Yamamoto
*Osaka University/Japan*

## ABSTRACT

*This scoping review explores the ethical challenges of using ChatGPT in higher education. By reviewing recent academic articles in English, Chinese, and Japanese, we aimed to provide a deep dive review and identify gaps in the literature. Drawing on Arksey & O'Malley's (2005) scoping review framework, we defined search terms and identified relevant publications from four databases in the three target languages. The research results showed that the majority of the papers were discussion papers, but there was some early empirical work. The ethical issues highlighted in these works mainly concern academic integrity, assessment issues, and data protection. Given the rapid deployment of generative artificial intelligence, it is imperative for educators to conduct more empirical studies to develop sound ethical policies for its use.*

55

## INTRODUCTION

Generative artificial intelligence (GAI), which is distinct from conventional pattern recognition technologies, is designed to create textual content based on human requirements. Beginning with the explosion of deep neural network technologies in 2012, leading technology companies have successively developed GAI models such as OpenAI's GPT-3 (Brown et al., 2020), Google's PaLM (Chowdhery et al., 2022), and Microsoft's Turing-NLG (Smith et al., 2022). The early versions of these products were found to potentially generate false, discriminatory, and harmful content. Consequently, researchers (Chung et al., 2022; Ouyang et al., 2022) have devoted considerable effort to aligning model outputs with human values. The launch of ChatGPT in November 2022 made the public aware that GAI was already capable of generating human-quality conversation, retrieving stored knowledge on demand, and achieving natural interaction with people as AI assistants. This kicked off the current frenzy of adapting the utilization of GAI to various fields, including education.

The advent of the ChatGPT marked a significant innovation in the realm of higher education, and it is rapidly extending its influence across multiple sectors, including teaching, learning, research, administration, and community engagement (UNESCO IESALC, 2023). The utility of ChatGPT within educational settings is multifaceted, spanning personalized learning pathways, curriculum enhancement, and the assessment of homework, exams, and essays (Huang, 2023; Kashiwamura, 2023; Ojha et al., 2023). By enhancing both educational practices and research methodologies, ChatGPT has emerged as a pivotal tool in advancing higher education's mission to foster learning and discovery (Farrokhnia et al., 2023).

Despite its numerous advantages, GAI studies have revealed potential risks associated with the generation of incorrect information (known as the 'hallucination' issue), biases (including race, nationality, and gender), and discriminatory content (Munn, 2023; Nozza et al., 2022). The existing body of literature on AI underscores concerns regarding fairness in application, attributing these to biases present in the outputs generated by language models (Benjamin, 2019; O'Neil, 2016). When GAI-generated outputs are used in education-related procedures, there is a possibility that problematic content, biases, and assumptions will be magnified, which may have negative consequences for learners, educators, researchers, and administrators.

This scoping review examines the ethical implications associated with the deployment of ChatGPT within the educational sector, with a specific emphasis on higher education. Through an analysis of academic articles in English, Chinese, and Japanese—languages in which we possess advanced proficiency—we endeavour to delineate the current landscape of research in this area. Our goal is to identify existing research gaps and outline potential directions for future investigation, thereby contributing to a comprehensive understanding of the ethical dimensions of using ChatGPT in educational contexts, particularly in teaching and learning, research or administration.

## RESEARCH METHOD

A scoping review is commonly used to identify key issues in a newly emerging field or one where there is not yet a substantial body of literature. It is "used to identify knowledge gaps, set research agendas, and identify implications for decision-making" (Tricco et al., 2016). In this study, we adopted Arksey and O'Malley's (2005) five-stage scoping review framework, which involves identifying the initial research questions and relevant studies, selecting the studies, charting the data, and collating, summarizing, and reporting the results.

**Identifying the relevant studies**

We limited our attention to articles focusing on the latest version of the GPT. We conducted the search in August 2023 and searched for articles that were published that year. We used the search terms "ChatGPT" or "Generative AI" coupled with "education" and "ethics" (see Table 1). To capture more solid evidence-based studies and discussions on this topic, we identified Scopus as the main database for our initial search. To include ongoing research, we also included the arXiv platform, which provides access to preprint articles. We included two other languages in which the authors had first or near-first language proficiency, Japanese or Chinese. To facilitate this, we conducted searches in the prominent databases CiNii (Japanese) and CNKI (Chinese). Along with the UK and USA, Japan and China are leading in AI development, making these languages good targets.

**Table 1: Final search terms and results by platform**

| Database | Search terms | Results |
|---|---|---|
| Scopus | (TITLE-ABS-KEY ("chatgpt" OR "generative AI") AND TITLE-ABS-KEY ("education" )) | 276 |

| | (TITLE-ABS-KEY ("chatgpt" OR "generative AI") AND TITLE-ABS-KEY ("education" AND "ethics")) | 27 |
|---|---|---|
| ArXiv | (TITLE-ABS-KEY ("chatgpt" OR "generative AI") AND TITLE-ABS-KEY ("education")) | 112 |
| | (TITLE-ABS-KEY ("chatgpt" OR "generative AI") AND TITLE-ABS-KEY ("education" AND "ethics")) | 24 |
| CiNii | (TITLE-ABS-KEY ("chatgpt" OR "生成 AI") AND TITLE-ABS-KEY ("教育")) | 23 |
| | (TITLE-ABS-KEY ("chatgpt" OR "生成 AI") AND TITLE-ABS-KEY ("教育"AND "課題")) | 4 |
| CNKI | (TITLE-ABS-KEY ("chatgpt" OR "生成 AI") AND TITLE-ABS-KEY ("教育")) | 198 |
| | (TITLE-ABS-KEY ("chatgpt" OR "生成 AI") AND TITLE-ABS-KEY ("教育" AND "伦理")) | 12 |

**Charting the data and collation**

The initial search yielded 609 results, of which 67 included education and ethical concerns. From these, we identified 26 articles meeting our inclusion criteria (Figure 1). All the articles were reviewed by two reviewers, and the third reviewer checked the findings.

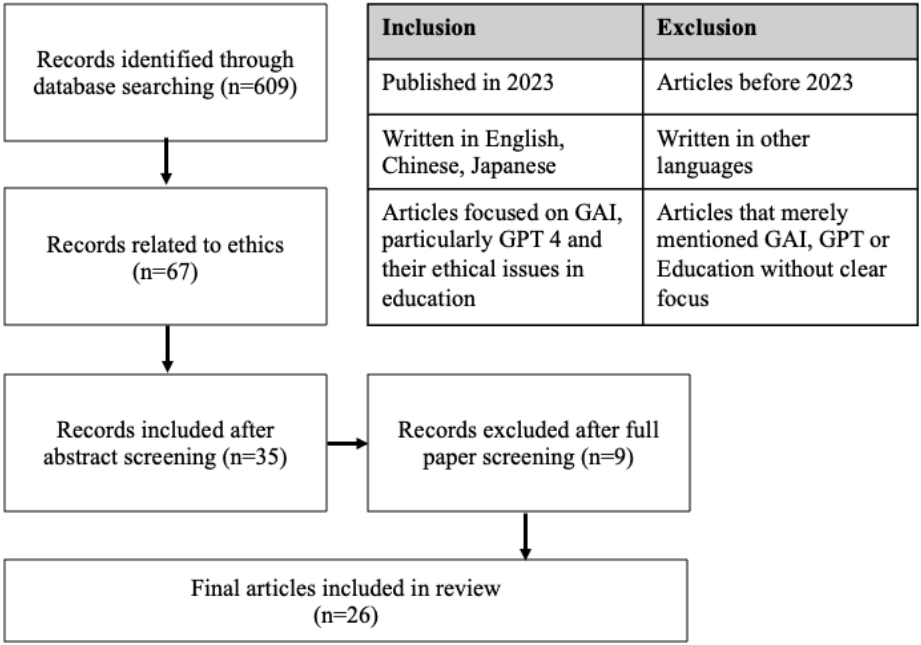

**Figure 1: Data extraction processes**

In our analysis of the ethical issues raised in the articles, we relied on the comprehensive research conducted by DeepMind (Weiginger et al., 2021), which offers a framework for assessing the ethical and social risks of harm that may arise from the deployment of language models (LMs) (Table 2).

**Table 2: Ethical and social risk areas**

| # | Areas | Description |
|---|---|---|
| 1 | Discrimination, Exclusion and Toxicity | AI models can harm by reinforcing discrimination, stereotypes, and biases, marginalizing individuals, promoting toxic language, and worsening disparities for disadvantaged groups. |
| 2 | Information Hazards | Leak of private data or sensitive information leaks. |

| 3 | Misinformation Harms | Providing false or misleading information, leading to less informed users and eroding trust in shared information. |
| 4 | Malicious Uses | Risks of using LMs for harm include enabling disinformation campaigns, personalized scams, fraud at scale, and the development of malicious computer code or weapon systems. |
| 5 | Human-Computer Interaction Harms | Users' overestimation of "human-like" AI capabilities may lead to unsafe usage, exploitation for manipulation, and perpetuation of stereotypes. |
| 6 | Automation, Access, and Environmental Harms | Unequal benefits and limited access to LMs can impact job quality, creative economy, and create global disparities in risks and rewards. |

## RESULTS

Most of the identified papers were in English (n=19), followed by Chinese (n=4) and Japanese (n=3). Among the English papers, ten were empirical studies, and nine were conceptual or discussion papers, with a predominant focus on their applications in fields such as healthcare and medical domains. In comparison, the number of Chinese and Japanese papers was much smaller. Among the four Chinese papers, all were general discussions about the application and predicted impact of the ChatGPT in education. There were only three Japanese articles and one that reported on initial research on students' practical experiences with ChatGPT specifications (Kondo et al., 2023). The other two papers discussed challenges for Japanese speakers in writing English academic papers and the use of ChatGPT for support and general teaching implications (Kashimura, 2023; Yanase, 2023).

      Overall, there has been little discussion specifically focusing on higher education. The majority of the papers (n=19) were generic, discussing ethical concerns in teaching (n=19) and learning (n=13) mostly from theoretical and conceptual perspectives without delving into specific levels of education. Papers that specifically focused on tertiary education (n=6) were concerned about overall pedagogical implications, particularly in medical education (n=2), faculty and students' perceptions (n=3), and research implications (n=1).

**Table 3: Articles reviewed**

| Authors | Language | Education level | Main area/Focus | Ethical concerns |
|---|---|---|---|---|
| **Database: Scopus** | | | | |
| Busch et al. | English | Tertiary | Teaching, learning, administration | 1,2, 3, 4, 5 |
| Chan | English | Tertiary | Teaching, learning | 2, 3, 5 |
| Curtis | English | Tertiary | Research | 3, 5 |
| da Silva | English | Generic | Research | 3, 5 |
| Dwivedi et al. | English | Generic | Research | 1, 2, 3, 4, 5, 6 |
| Fischer | English | Generic | Administration | 1, 2, 3, 4, 5 |
| Krüger et al. | English | Generic | Teaching, learning, research | 1, 2, 3, 5 |
| Lim et al. | English | Generic | Teaching | 1, 2, 3 |
| Masters (a) | English | Generic | Teaching, administration | 1, 2, 3, 4, 5 |
| Masters (b) | English | Generic | Research | 3, 4 |
| O'Connor & ChatGPT | English | Generic | Teaching, learning, research | 3, 5 |
| Tlili et al. | English | Generic | Teaching, learning | 1, 2, 3, 4, 5, |
| Zumsteg & Junn | English | Tertiary | Teaching, learning | 3, 4, 5 |
| **Database: arXiv** | | | | |
| Chan & Hu | English | Tertiary | Teaching, learning | 1, 2, 3, 4, 5 |
| Latif et al. | English | Generic | Teaching | 1, 2, 3, 4, 5 |
| Li et al. | English | Generic | Teaching, learning, research | 1, 2, 3, 4, 5 |
| Ojha et al. | English | Generic | Teaching | 4, 5 |
| Sharma et al. | English | Generic | Administration | 3, 4 |
| Sharples | English | Generic | Teaching, learning | 3, 5 |
| **Database: CiNii** | | | | |

| Kashimura | Japanese | Generic | Teaching | 1, 2, 3 |
| Kondo et al. | Japanese | Secondary | Teaching, learning | 3, 5 |
| Yanase | Japanese | Generic | Research | 3, 5 |
| **Database: CNKI** | | | | |
| Huang | Chinese | Generic | Teaching, learning | 1, 3, 5 |
| Song & Lin | Chinese | Generic | Teaching | 2, 3, 5 |
| Xun | Chinese | Tertiary | Teaching, learning | 1, 3, 4, 5 |
| Zhu & Yang | Chinese | Generic | Teaching, learning | 1, 2, 3, 5 |

In terms of the focus of ethical issues, the majority of papers concerned with #3 misinformation harms (n=25), including academic integrity, cheating and other assessment issues, and the role of users in identifying and clarifying information and/or #5 human-computer interaction-related harms (n=24), such as addiction, dependence, and cognitive overload. To illustrate this further, we divided the papers into four themes concerning teaching, learning, research, and administration.

In this section, we consolidate the principal concerns and discussion points from the literature regarding the ethical implications of the ChatGPT in higher education.

**Teaching**

The literature concerning teaching mostly addressed misinformation harms (n=19), followed by human-computer interaction harms (n=18). This includes pedagogical implications for incorporating AI in university teaching, such as during assessments. Research by Latif et al. (2023) highlighted the risk of AI perpetuating existing societal biases related to gender and nationality, sourced from training data, which could adversely affect the fairness and integrity of educational applications. This underscores the importance of cautious reliance on AI for evaluations, as it may not always accurately reflect students' abilities (Busch et al., 2023; Curtis, 2023; Song & Lin, 2023).

The integration of AI in education also prompts a reevaluation of the educator-student dynamic. Concerns about a potential overdependence on AI-generated content, as discussed by Sharples (2023), highlight the risk of undermining the traditional roles of educators, potentially detracting from their unique contributions to crafting engaging and innovative lesson plans and learning activities. This calls for a balanced approach to integrating AI in education, ensuring that it complements rather than diminishes the value of human instruction.

ChatGPT facilitates tailored learning experiences and bolsters student support, language tutoring, content generation, and career guidance (Kooli, 2023; Lim et al., 2023). However, to design new programs or to provide personalized teaching, universities need to collect and process vast amounts of student data, often without students' consent. This raises substantial concerns regarding data privacy and security, underscoring the need for stringent data protection measures to safeguard sensitive information against unauthorized use (Chan, 2023; Masters, 2023a).

**Learning**

Similar to teaching, the literature on "learning" concerned misinformation harm and human-computer interaction harm (n=12). This includes concern for a potential increase in plagiarism and cheating among students who might rely on GAI-generated content for essays and exams, thereby compromising the authenticity of their work (Li et al., 2023; Zhu & Yang, 2023). This overreliance on ChatGPT may lead to a decline in students' sense of responsibility and commitment to academic integrity (Ojha et al., 2023). The overuse of ChatGPT may adversely affect students' critical thinking skills by leading to heavy dependence on AI-generated content, which can diminish their ability to independently analyze and evaluate information (Tlili et al., 2023).

Another significant issue raised in the literature pertains to the risk of misinformation being propagated due to the highly persuasive and convincing nature of AI-generated content (Chan & Hu, 2023; Latif et al., 2023; Li et al., 2023). This can lead to potential bias or manipulation of information presented to students.

Reliance on AI interactions for academic or social purposes might diminish face-to-face interactions, potentially hindering the development of essential social skills among students. A striking balance between AI and human interactions is crucial to fostering a well-rounded educational experience (Kondo et al., 2023; Zumsteg & Junn, 2023).

ChatGPT might inadvertently produce content that inaccurately or inappropriately represents certain cultural or identity groups, highlighting the need for ongoing refinement and sensitivity in AI language model development (Busch et al., 2023).

**Research**

ChatGPT offers efficient dataset analysis, automated code generation, comprehensive literature reviews, and streamlined experimental design processes (Dwivedi et al., 2023; Li et al., 2023). These capabilities underscore its potential to accelerate research discovery and innovation. However, the literature has discussed the potential generation of misleading information (n=8) or the exploitation and perpetuation of stereotypes. For example, the attribution of fake references to AI-generated content leads to misinformation and a decline in trust in academic sources (Curtis, 2023).

The integration of AI in academic publishing poses the risk of displacing human authors and undermining the value of their expertise, potentially impacting the credibility of research. The joint authorship of editorial pieces such as the one addressed in O'Connor and ChatGPT (2023) challenges the established core values related to human-based authorship in academic publishing (da Silva, 2023).

Some conferences permit the use of ChatGPT for writing papers, but only when ChatGPT itself is the subject of empirical research (e.g., ICML, 2023). On the other hand, some research communities, such as the Association for Computational Linguistics (ACL, 2023), allow the use of the ChatGPT based on specific guidelines.

**Administration**

Only four literature items discussed the administrative aspect of AI application in higher education institutions. Four studies touched on malicious use and misinformation harm. From a positive perspective, ChatGPT can significantly reduce the time spent on human administrative tasks, such as responding to queries from applicants and assisting students in course enrollment (UNESCO IESALC, 2023). This efficiency in handling administrative duties not only optimizes operational processes but also allows staff to dedicate more time to tasks that require human touch, further enhancing educational infrastructure.

However, there are concerns surrounding the equitable, reliable, and transparent use of the ChatGPT. Utilizing ChatGPT in admissions processes can potentially introduce biases, especially if the AI model is trained on historical data that reflects past inequalities (Fischer, 2023; Sharma et al., 2023). To ensure fairness, transparency, and accountability, it is essential to provide applicants with clear explanations of how AI was employed to assess their applications and the specific factors that contributed to their acceptance or rejection.

Additionally, the use of AI algorithms in admissions decisions carries the risk of inadvertently favoring applicants with certain characteristics or backgrounds, potentially impacting diversity and inclusion efforts within the university (Busch et al., 2023; Fischer, 2023). Data privacy and security are also paramount considerations. To avoid unintentional discrimination, institutions should actively assess and address any biases in the AI model's training data and decision-making process, striving to provide equal opportunities for all applicants.

# CONCLUSION

We focused on articles written in a very short period, the first 7 months of 2023, but covered literature written in English, Chinese, and Japanese. Given that the Chat GPT is trained on English-centric data (Brown et al., 2020), it is important to gain insights into discussions of non-English-speaking AI technologically advanced countries. However, our review revealed that a few academic and research studies were published in languages other than English, particularly in Chinese and Japanese.

Our scoping review showed that there are already publications that are considering the ethical implications of the GAI, especially the ChatGPT, in education generally, and some have focused on higher education. The majority of papers are discussion works, but there is some early empirical work. The ethical issues highlighted in these works mainly concern academic integrity, assessment issues, and data protection.

The increased use of GAI by learners raises issues related to academic integrity, definitions of authorship, assessment methods, and other pedagogical implications (Li et al., 2023; Ojha et al., 2023). It also affects how researchers conduct their studies and generate their outputs, as well as how decisions are made in admissions, hiring, or how educational institutions are managed and run (Fischer, 2023; Master, 2023a; Sharma et al., 2023). Furthermore, the increasing integration of AI in education has even raised questions about the continued relevance of traditional brick-and-mortar educational institutions (Sharples, 2023). Therefore, it is crucial to further discuss and assess the ethical implications of implementing the ChatGPT within educational institutions, especially concerning its use in teaching and learning, research or administration.

Our analysis highlights the urgency of addressing ethical issues surrounding the use of the GAI/ChatGPT in education. Collaboration among stakeholders is essential for establishing clear guidelines, protecting student privacy, and promoting responsible AI use. By doing so, AI can enhance education and research without compromising fundamental principles.

**MING LI (PhD)** is Associate Professor at the Institute for Transdisciplinary Graduate Degree Programs, Osaka University, Japan. Her major academic fields include internationalization of higher education, international student mobility, and interdisciplinary education. Email:li.ming.itgp@osaka-u.ac.jp

**ARIUNAA ENKHTUR (PhD)** is Specially Appointed Assistant Professor at the Center for Global Initiatives, Osaka University, Japan. Her research interests include internationalization of higher education, academic mobility, virtual student mobility, transnational higher education, international education and development, teaching and learning.

**FEI CHENG (PhD)** is a program-based Junior Associate Professor/Senior Lecture at Kyoto University, Japan. He received his PhD in Informatics from NARA Institute of Science and Technology in 2018. His research interests include information extraction, numerical reasoning, large language models, and a broad range of natural language processing research.

**BEVERLEY ANNE YAMAMOTO (PhD)** is currently serving as an executive vice president of international affairs (education) and concurrently holding the position of professor in the Graduate School of Human Sciences. Her research interests include the internationalization of education, stakeholder engagement, including in relation to healthcare AI, and school-based health promotion.